\documentclass{article}
\usepackage{iclr2026_conference,times}  

\usepackage{amsmath,amsfonts,bm}

\def\eqref#1{equation~\ref{#1}}

\def\1{\bm{1}}

\DeclareMathAlphabet{\mathsfit}{\encodingdefault}{\sfdefault}{m}{sl}
\SetMathAlphabet{\mathsfit}{bold}{\encodingdefault}{\sfdefault}{bx}{n}

\usepackage{hyperref}
\usepackage{url}
\usepackage{booktabs}
\usepackage{multirow}
\usepackage{graphicx}
\usepackage{xcolor}
\usepackage{placeins}
\usepackage[table]{xcolor}
\definecolor{lightgray}{gray}{0.9}
\definecolor{lightyellow}{RGB}{255, 255, 200}

\title{Finish First, Perfect Later: \\ 
Test-time Token-Level Cross-Validation for Diffusion Large Language Models}

\author{
Runchu Tian\textsuperscript{1}\thanks{Equal contribution.}
\quad Junxia Cui\textsuperscript{2}\footnotemark[1]
\quad Xueqiang Xu\textsuperscript{1}
\quad Feng Yao\textsuperscript{2}
\quad Jingbo Shang\textsuperscript{2}\\
\ \textsuperscript{1}University of Illinois Urbana-Champaign
\quad
\textsuperscript{2}University of California San Diego\\
\ \texttt{runchut2@illinois.edu, jucui@ucsd.edu, jshang@ucsd.edu}
}

\iclrfinalcopy  

\begin{document}
\maketitle

\begin{abstract}







Diffusion large language models (dLLMs) have recently emerged as a promising alternative to autoregressive (AR) models, offering advantages such as accelerated parallel decoding and bidirectional context modeling.
However, the vanilla decoding strategy in discrete dLLMs suffers from a critical limitation: once a token is accepted, it can no longer be revised in subsequent steps. 
As a result, early mistakes persist across iterations, harming both intermediate predictions and final output quality.
To address this issue, we propose \textsc{Tolerator} (\textbf{To}ken-\textbf{Le}vel C\textbf{r}oss-V\textbf{a}lida\textbf{t}i\textbf{o}n \textbf{R}efinement), a training-free decoding strategy that leverages cross-validation among predicted tokens. 
Unlike existing methods that follow a single progressive unmasking procedure, \textsc{Tolerator} introduces a two-stage process: (i) sequence fill-up and (ii) iterative refinement by remasking and decoding a subset of tokens while treating the remaining as context. 
This design enables previously accepted tokens to be reconsidered and corrected when necessary, leading to more reliable diffusion decoding outputs.
We evaluate \textsc{Tolerator} on five standard benchmarks covering language understanding, code generation, and mathematics. 
Experiments show that our method achieves consistent improvements over the baselines under the same computational budget.
These findings suggest that decoding algorithms are crucial to realizing the full potential of diffusion large language models. Code and data are publicly \href{https://github.com/Rachum-thu/tolerator}{available}.
\end{abstract}

\section{Introduction}

Large language models (LLMs)~\citep{chowdhery2022palmscalinglanguagemodeling, hurst2024gpt, comanici2025gemini} have driven remarkable progress across diverse NLP domains~\citep{zhao2023survey, minaee2024large}. 
The dominant architecture behind these advances is the autoregressive (AR) transformer~\citep{NIPS2017_3f5ee243}. While highly effective, AR decoding is inherently sequential, creating a fundamental bottleneck that limits generation parallelism~\citep{10.5555/3692070.3692631, xia-etal-2024-unlocking}.

To address this, diffusion language models~\citep{NEURIPS2021_958c5305, Diffusion-LM} have emerged as a powerful alternative, generating sequences through iterative denoising with bidirectional attention and parallel token predictions. This paradigm offers distinct advantages over AR models~\citep{li2025surveydiffusionlanguagemodels}, including accelerated inference, stronger global coherence, and controllable quality–speed trade-offs. Recent progress~\citep{labs2025mercuryultrafastlanguagemodels, nie2025largelanguagediffusionmodels, ye2025dream7bdiffusionlarge} has further demonstrated the practicality and competitiveness of discrete diffusion large language models (dLLMs). Commercial dLLMs such as Mercury Coder~\citep{labs2025mercuryultrafastlanguagemodels} and Gemini Diffusion~\citep{gemini_diffusion} claim to match the performance of autoregressive LLMs~\citep{hurst2024gpt, jambateam2024jamba15hybridtransformermambamodels} while achieving up to $10\times$ faster inference on tasks like code generation~\citep{chen2021codex, austin2021program}.

\begin{figure*}[t]
  \centering
  \includegraphics[width=1.00\linewidth]{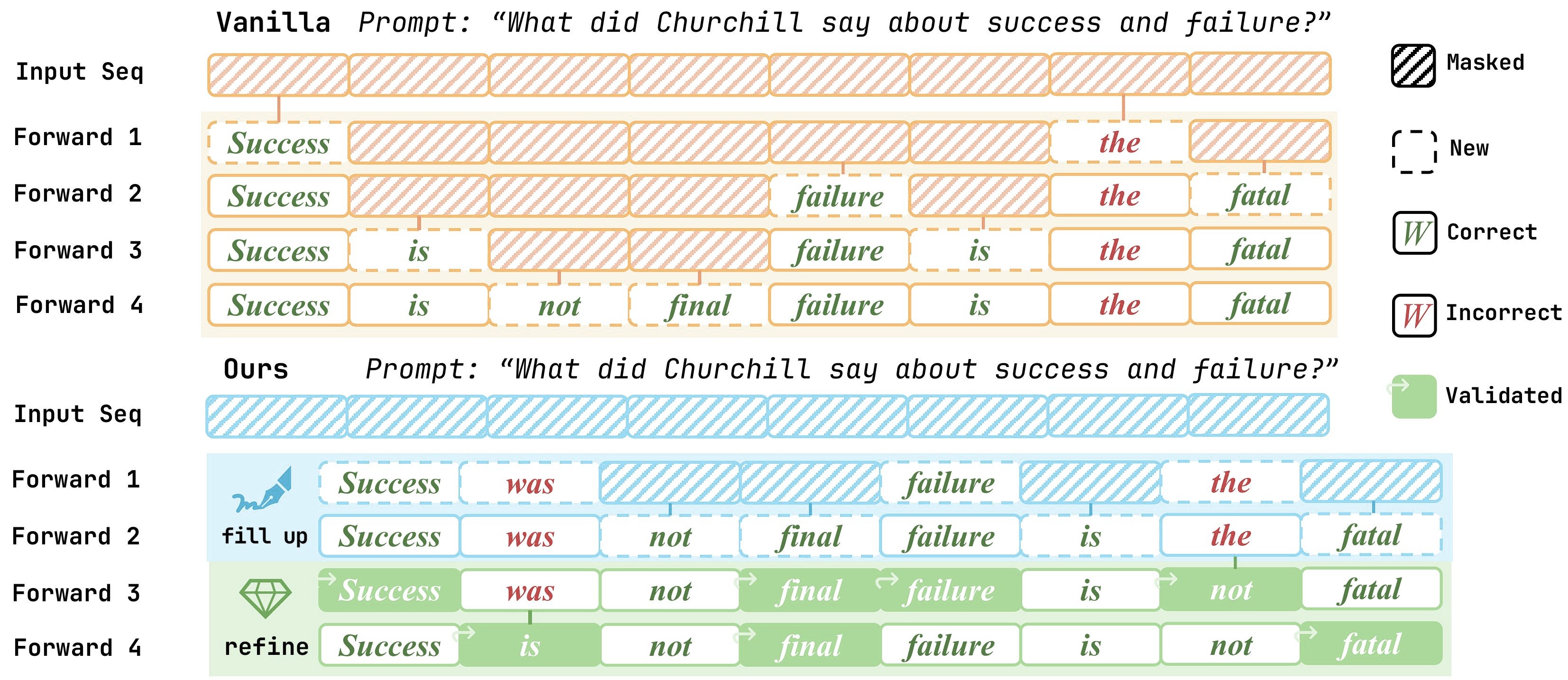}
  \label{fig:overview}
  \vspace{-0.5em}
  \caption{\textbf{Overview of \textsc{Tolerator}}. Compared to the vanilla decoding strategy, we first fill the masked tokens with high parallelism and then iteratively refine the draft through token-level cross-validation. Here, cross-validation means tokens alternately act as the target and the context of prediction. This process allows previously accepted tokens to be revisited and corrected when necessary.}
  \vspace{-1.5em}

\end{figure*}

Despite recent advances, current discrete dLLM decoding strategies~\citep{israel2025acceleratingdiffusionllmsadaptive, yu2025dimplediscretediffusionmultimodal, wu2025fastdllmtrainingfreeaccelerationdiffusion} suffer from a critical limitation: once a token is accepted, it is typically fixed and cannot be modified in later steps~\citep{wang2025remaskingdiscretediffusionmodels,vonrütte2025generalizedinterpolatingdiscretediffusion}. For instance, in two widely adopted open-source dLLMs, LLaDA~\citep{nie2025largelanguagediffusionmodels} and Dream~\citep{ye2025dream7bdiffusionlarge}, a token is considered \emph{accepted} if, at a specific iteration, it is unmasked and no longer remasked, as illustrated in Figure~\ref{fig:overview}.
Once accepted, it will serve as fixed context for all future predictions. This causes early mistakes to persist and propagate throughout the generation process~\citep{wang2025remaskingdiscretediffusionmodels, vonrütte2025generalizedinterpolatingdiscretediffusion}.

There have been some early explorations on this issue. ReMDM~\citep{wang2025remaskingdiscretediffusionmodels} introduces a sampler that applies a stochastic backward remasking process for predicted tokens. RCR~\citep{he2025mdpoovercomingtraininginferencedivide} tracks each token’s running max confidence and remasks persistently low-confidence tokens. GIDD~\citep{vonrütte2025generalizedinterpolatingdiscretediffusion} trains diffusion models with a mixing schedule that interpolates between data and noise distributions to enable the remasking of predicted tokens. While these works demonstrate the significance of dLLM decoding strategies, their improvements have not achieved ideal performance on general tasks, so the challenge remains an open problem.

To further bridge this gap, we propose \textsc{Tolerator} (\textbf{To}ken-\textbf{Le}vel C\textbf{r}oss-V\textbf{a}lida\textbf{t}i\textbf{o}n \textbf{R}efinement), a test-time dLLM decoding method that explicitly separates generation into two stages: \textbf{fill-up} and \textbf{refinement}. In the first stage, we fill up the masked tokens following vanilla dLLM decoding strategy. In the second stage, we iteratively refine this draft by remasking and decoding subsets of tokens while using the remaining as context, so that predictions are revised by cross-validating against one another. 
Our approach differs from existing strategies which perform refinement within the ongoing generation process. By explicitly decoupling fill-up and refinement into two separate phases, \textsc{Tolerator} enables a more thorough form of token-level error correction than prior methods.

We evaluate \textsc{Tolerator} on five standard benchmarks across language understanding (TriviaQA~\citep{joshi-etal-2017-triviaqa}, GPQA~\citep{rein2024gpqa}), code generation (MBPP~\citep{austin2021program}, HumanEval~\citep{chen2021codex}), and mathematics (GSM8K~\citep{cobbe2021trainingverifierssolvemath}). We use vanilla decoding, ReMDM~\citep{wang2025remaskingdiscretediffusionmodels}, and RCR~\citep{he2025mdpoovercomingtraininginferencedivide} as baselines. Experimental results show that, under the same computational cost measured by the number of forward steps, \textsc{Tolerator} achieves noticeable and consistent improvements over the baselines (relatively improves 17.9\% for Dream~\citep{ye2025dream7bdiffusionlarge} and 15.3\% on LLaDA~\citep{nie2025largelanguagediffusionmodels}). Overall, our findings indicate that decoding strategy is not merely an implementation detail, but a crucial factor that substantially influences the performance of dLLMs.





\section{Related Work}\label{sec:related}

\subsection{From Autoregression to Diffusion}
Modern natural language generation~\citep{hendrycks2020measuring, suzgun-etal-2023-challenging, rein2024gpqa} has been dominated by the autoregressive (AR) model architectures like GPT~\citep{brown2020languagemodelsfewshotlearners} and LLaMA~\citep{touvron2023llamaopenefficientfoundation}. Despite its empirical success, AR models introduce a fundamental bottleneck: generation is inherently sequential, limiting decoding parallelism~\citep{10.24963/ijcai.2023/750, zou2023surveydiffusionmodelsnatural}. To address this limitation, diffusion language models~\citep{NEURIPS2021_958c5305, Diffusion-LM} have emerged as a promising alternative~\citep{li2025surveydiffusionlanguagemodels}. By reversing a noising process over multiple steps, diffusion language models generate tokens in parallel~\citep{labs2025mercuryultrafastlanguagemodels} while leveraging full bidirectional attention~\citep{nie2025largelanguagediffusionmodels, ye2025dream7bdiffusionlarge}.

Existing diffusion language models can be classified into three main categories depending on how the diffusion process is applied. Early \emph{continuous diffusion language models}~\citep{Diffusion-LM, strudel2022selfconditionedembeddingdiffusiontext, karimi-mahabadi-etal-2024-tess, lovelace2023latent, dieleman2022continuous} denoised \emph{embeddings} before mapping them back to tokens. However, this paradigm struggles with issues like optimization and has largely been replaced by discrete diffusion language models. \emph{Discrete diffusion language models}~\citep{NEURIPS2021_958c5305, he-etal-2023-diffusionbert} define diffusion directly in \emph{token} space, and further scale up model parameter size~\citep{gong2025scaling}, achieving the state-of-the-art with open-source models like Dream~\citep{ye2025dream7bdiffusionlarge} and LLaDA~\citep{nie2025largelanguagediffusionmodels}. A third line integrates AR philosophy with diffusion, including block-wise or multi-level scheduling~\citep{han-etal-2023-ssd, wu2023ar} and the reintroduction of sequential dependency while retaining diffusion-style refinement~\citep{arriola2025block, huang2025ctrldiffboostinglargediffusion}.

\subsection{Training and Inference Strategies in Diffusion Language Models}
Beyond architectural explorations, another line of work studies how to effectively train diffusion LMs. 
Large-scale instruction tuning~\citep{ye2025dream7bdiffusionlarge, nie2025largelanguagediffusionmodels} has demonstrated that diffusion models can achieve general capabilities comparable to autoregressive LLMs. 
Researchers also explore refinements of the training objective: simplified masked losses~\citep{shi2024simplified, sahoo2024simple}, likelihood-based formulations~\citep{gulrajani2023likelihood}, and variants that enhance generation robustness and reasoning~\citep{vonrütte2025generalizedinterpolatingdiscretediffusion, ye2025autoregressiondiscretediffusioncomplex}. 
Another direction focuses on adapting reinforcement learning to diffusion, either to strengthen reasoning~\citep{huang2025ctrldiffboostinglargediffusion, NEURIPS2024_be30024e, zhao2025d1} or for preference optimization~\citep{zhu2025llada}. 

Decoding is another key bottleneck for diffusion language models: parallel generation improves efficiency but often degrades quality. Adaptive Parallel Decoding (APD)~\citep{israel2025acceleratingdiffusionllmsadaptive} mitigates this trade-off by adjusting the degree of parallelism with an auxiliary autoregressive verifier, while dilated~\citep{luxembourg2025planspeeddilatedscheduling} scheduling further accelerates inference. At the same time, KV-caching~\citep{ma2025dkvcachecachediffusionlanguage, wu2025fastdllmtrainingfreeaccelerationdiffusion} and autoregressive-guided unmasking~\citep{hu2025acceleratingdiffusionlanguagemodel} are applied to further accelerate dLLMs. Recent works~\citep{li2025fixedtrainingfreevariablelengthdenoising, kim2025anyorderflexiblelengthmasked} also address flexibility by extending diffusion to variable-length and token insertion.

\subsection{Error Correction in Diffusion Decoding}

It is often claimed that vanilla discrete diffusion language models possess an inherent ability for error correction, since each position is repeatedly predicted as the context evolves over iterations~\citep{wang2025remaskingdiscretediffusionmodels, gemini_diffusion}. However, this view is incomplete: once a token is accepted, it becomes fixed and cannot be revised. For example, LLaDA~\citep{nie2025largelanguagediffusionmodels} and Dream~\citep{ye2025dream7bdiffusionlarge} decide at every iteration whether a token should be further remasked; if it is not, the token is considered accepted and remains unchanged thereafter. As a result, any early mistake will persist and propagate through subsequent steps, limiting the reliability of diffusion generation.

Several methods have sought to address this limitation. ReMDM~\citep{wang2025remaskingdiscretediffusionmodels} introduces a probabilistic remasking process that allows already revealed tokens to be re-predicted. RCR~\citep{he2025mdpoovercomingtraininginferencedivide} proposes a simple confidence-based strategy that remasks uncertain tokens during inference. GIDD~\citep{vonrütte2025generalizedinterpolatingdiscretediffusion} modifies the corruption process with hybrid noise during training. While these approaches demonstrate the feasibility of token revision, their empirical gains remain relatively modest on general tasks or they require additional training, leaving the core problem unresolved. In contrast, our approach departs from prior work by explicitly decoupling fill-up and refinement. We first generate a draft following vanilla diffusion decoding, and then apply a targeted refinement stage that revisits the accepted tokens according to a cross-validation principle. This separation not only makes error correction conceptually more systematic but also delivers markedly stronger empirical gains.

\section{Methodology}

\subsection{Preliminaries}

\paragraph{Decoding in Discrete dLLMs. }
\label{sec: dllm-decode}

We consider the decoding process of discrete diffusion large language models~\citep{ye2025dream7bdiffusionlarge, nie2025largelanguagediffusionmodels}. Specifically, let \( x^{(t)}_i \in \mathcal{V} \) denote the token at position \( i \in \{1,\dots,L\} \) and time step \( t \in \{0,\dots,T\} \), where \( \mathcal{V} \) is the vocabulary, \( L \) is the sequence length, and \( T \) is the total number of forward steps. At inference time, the sequence is initialized with
\[
x^{(0)}\;=\; \bigl( c_1,\dots ,c_m,\underbrace{\text{[\texttt{MASK}]}_{m+1},\dots ,\text{[\texttt{MASK}]}_{L}}_{L-m} \bigr) \;\in\;\mathcal{V}^{L},
\]
where \( c_{1} \) to \( c_{m} \) are prompt tokens and the remaining \( L - m \) positions are masked tokens. At each time step, discrete diffusion large language models output the logits of all masked tokens and decode them by sampling, where \( y^{(t)}_i \sim p_{\theta}(\cdot \mid x^{(t)}, t) \), and \( p_{\theta} \) is the conditional distribution parameterized by the dLLMs. A deterministic rule then decides whether to accept or remask each decoded token.
Specifically, the next sequence is constructed as
\[
x^{(t+1)}_i =
\begin{cases}
y^{(t)}_i, & \text{accepted},\\[6pt]
\text{[\texttt{MASK}]}, & \text{remasked},
\end{cases}
\quad \text{for } i \notin \mathcal{I}_t,
\qquad
x^{(t+1)}_j = x^{(t)}_j \quad \text{for } j \in \mathcal{I}_t.
\]

where \( \mathcal{I}_t \subseteq \{1,\dots,L\} \) is the index set of tokens already accepted at step \( t \). In the vanilla setup, each step accepts approximately \( \lfloor L/T \rfloor \) tokens, which are selected based on criteria like model confidence or entropy. Different dLLMs may adopt alternative decoding strategies. For example, semi-autoregressive decoding~\citep{nie2025largelanguagediffusionmodels} only proceeds to the next block once all tokens in the current block have been accepted. Our study focuses on the vanilla setup, as it is widely adopted in existing dLLMs.

\paragraph{Limitations of Conventional Discrete dLLM Decoding.}
In the conventional decoding setup, masked positions are iteratively unmasked and decoded, while accepted tokens become fixed and remain unchanged. Formally, once a position index \( i \) enters the visible set \( \mathcal{I}_t \), we have \( i \in \mathcal{I}_{t'} \) and \( x^{(t')}_i = x^{(t)}_i \) for all \( t' > t \). As a result, an early error at position \( j \in \mathcal{I}_t \) is permanently preserved and enters the context for all future predictions \( p_\theta(x^{(t')}_i \mid x^{(t'-1)}, t'-1) \), where \( i \notin \mathcal{I}_{t'-1} \). Such errors cannot be revised and may propagate through the decoding process as persistent noise, ultimately degrading the quality of the generated sequence.

\subsection{Method Overview}
To overcome this limitation, we propose \textsc{Tolerator} (\textbf{To}ken-\textbf{Le}vel C\textbf{r}oss-V\textbf{a}lida\textbf{t}i\textbf{o}n \textbf{R}efinement), which moves beyond the traditional view of decoding as a single, progressive unmasking trajectory, and instead reframes it as a two-stage process of \textbf{fill-up} and \textbf{refinement}. 

\paragraph{Stage~I (Sequence Fill-Up).} 
In the fill-up stage, the model produces a coarse draft by filling masked positions following the vanilla dLLM decoding strategy, providing a complete but potentially imperfect hypothesis of the output. 

\textbf{Stage~II (Cross-Validation Refinement).} 
In the refinement stage, our iterative procedure follows a token-level cross-validation principle, where tokens alternately act as validator and as validation targets. This alternating role improves the overall consistency of generated sequence.

Overall, this design offers a training-free, model-agnostic solution to the challenge of irreversible early errors and their propagation in the decoding process.

\subsection{Sequence Fill-Up}

The sequence fill-up stage is largely based on the vanilla dLLM decoding procedure described in Section~\ref{sec: dllm-decode}. However, to facilitate the refinement stage, we introduce one modification: the logit penalty on the End-of-Text (EoT) token.


\paragraph{EoT Penalty.}
Since the refinement stage can correct errors, we prefer longer and more informative drafts to overly short completions. To this end, we apply an \emph{EoT penalty}~\citep{bai-etal-2021-semantics, laban-etal-2020-summary}, which discourages the generation of EoT tokens in the fill-up stage. Concretely, we scale down the logit of the EoT token by a factor \(\lambda_{\text{eot}} > 1\) before softmax. While this adjustment does not directly improve draft quality, it effectively prevents early termination and produces drafts that are better suited for subsequent refinement. Formally, let \(z_v\) be the unnormalized logit for token \(v\) at position \(i\) and time step \(t\). The penalized distribution is
\[
\tilde{p}_\theta(v \mid x^{(t)}, t) \propto 
\begin{cases}
\exp(z_v) / \lambda_{\text{eot}}, & \text{if } v = \texttt{[EoT]} \\
\exp(z_v), & \text{otherwise.}
\end{cases}
\]

Finally, the fill-up stage produces a sequence consisting of the prompt tokens and model predictions for previously masked positions:
\[
x^{(\rho T)} = \bigl( c_1,\dots,c_m,\, x^{(\rho T)}_{m+1}, \dots, x^{(\rho T)}_L \bigr) \in \mathcal{V}^L,
\]
where \( x^{(\rho T)}_i \ne \texttt{[MASK]} \) for all \( i > m \). Here \(\rho \in (0,1)\) controls the split between the two stages.

\subsection{Cross-Validation Refinement}

The refinement stage corrects errors in the draft with a token-level cross-validation principle, where tokens alternately act as validator and as validation targets. In each iteration, a different subset of tokens is \emph{sampled}, \emph{remasked} and \emph{decoded} conditioned on the preserved context, progressively reducing mistakes and improving coherence.

\paragraph{Iterative Refinement.}
At each iteration \(k\), we remask a random subset \(S^{(k)} \subseteq \{m+1,\dots,L\}\) of non-prompt positions, sampled at rate \(\gamma_k\) so that \(|S^{(k)}| = \lfloor \gamma_k (L-m) \rfloor\).

$$
x^{(k)}_i =
\begin{cases}
\texttt{[MASK]}, & i \in S^{(k)} \\
x^{(k)}_i, & \text{otherwise.}
\end{cases}
$$

The sequence for the next iteration is then obtained by predicting the masked tokens:

$$
x^{(k+1)}_i =
\begin{cases}
y^{(k)}_i, & i \in S^{(k)} \\
x^{(k)}_i, & \text{otherwise,}
\end{cases}
\quad \text{where } y^{(k)}_i \sim p_\theta(\cdot \mid x^{(k)}, k).
$$

In each iteration, a subset of generated tokens is held fixed as context, while the remaining tokens are remasked and decoded to better align with the fixed context tokens. Iterating this process gradually improves the coherence of the entire sequence.

\paragraph{Annealed Refinement Rate.}
To improve the stability of refinement steps, we anneal the refinement rate $\gamma_k$ over time. Higher refinement rates in early iterations encourage broader corrections of initial errors, while lower rates in later iterations help stabilize the predictions. We adopt a cosine annealing schedule with both upper and lower bounds:

$$
\gamma_k \;=\; \gamma_{\min} + \tfrac{1}{2}(\gamma_{\max} - \gamma_{\min})\left(1 + \cos\! \ \bigl(\tfrac{\pi k}{K}\bigr)\right),
$$

where $k$ is the current refinement iteration and $K$ is the total number of refinement steps.

\section{Experiment}

\subsection{Experimental Setup}

\paragraph{Models.}
Following previous studies~\citep{ma2025dkvcachecachediffusionlanguage, israel2025acceleratingdiffusionllmsadaptive, wu2025fastdllmtrainingfreeaccelerationdiffusion, he2025mdpoovercomingtraininginferencedivide}, we evaluate our method on two representative open-source discrete dLLMs: \textbf{Dream-v0-Instruct-7B}~\citep{ye2025dream7bdiffusionlarge} and \textbf{LLaDA-8B-Instruct}~\citep{nie2025largelanguagediffusionmodels}. Both are state-of-the-art representatives of open-source discrete diffusion large language models. 

\paragraph{Datasets \& Metrics.}
To assess the general effectiveness of our method, we evaluate it on three representative tasks with five standard benchmarks: (i) language understanding with  \textbf{TriviaQA}~\citep{joshi-etal-2017-triviaqa} and \textbf{GPQA}~\citep{rein2024gpqa}, (ii) code generation with \textbf{HumanEval}~\citep{chen2021codex} and \textbf{MBPP}~\citep{austin2021program}, and (iii) mathematics with \textbf{GSM8K}~\citep{cobbe2021trainingverifierssolvemath}. We report accuracy for TriviaQA, GPQA, GSM8K and pass@1 for HumanEval and MBPP.  

\paragraph{Baselines.}
We compare our method against the vanilla decoding strategy and two training-free baselines that propose to revise the accepted tokens. (i) \textbf{Vanilla} follows the standard dLLM decoding procedure: once a token is accepted, it remains fixed throughout the generation process and cannot be revised. (ii) \textbf{ReMDM}~\citep{wang2025remaskingdiscretediffusionmodels} introduces a stochastic sampler that applies a backward remasking process for predicted tokens. (iii) \textbf{RCR}~\citep{he2025mdpoovercomingtraininginferencedivide} records each token’s running max confidence and remasks persistently low-confidence tokens. 


\paragraph{Configurations.}

For fairness, all methods are evaluated with the same dLLM backbones in the zero-shot setting. We equalize the computational cost between baselines and our method by allocating the same total forward-step budget. Specifically, for our method, we set the allocation ratio \(\rho\) between sequence fill-up and refinement to 0.5. However, it is worth noting that our method itself has no restriction on how steps are allocated. This constraint is introduced solely to ensure a fair comparison.

For our method, we adopt a cosine annealing scheduler for the refinement rate with \(\gamma_{\max} = 0.8\) and \(\gamma_{\min} = 0.4\), and increase the EoT penalty \(\lambda_{\text{eot}}\) from 1.0 to 1.3 as the number of forward steps $T$ grows. For baselines, we use the recommended hyperparameters for ReMDM (\(t_{\text{on}}=0.55\), \(t_{\text{off}}=0.05\), \(\alpha_{\text{on}}=0.9\)) and use the linear remasking scheduling function for RCR, which is reported to be optimal~\citep{he2025mdpoovercomingtraininginferencedivide}.  

We follow the default prompts from the LM-Eval framework~\citep{eval-harness} and fix the generation length \(L\) at 256. The total number of forward steps $T$ varies from 4 to 256 in powers of two, covering the scenarios from highly parallel to fully sequential decoding. All experiments are run on 8 NVIDIA H200 GPUs, and each data point is evaluated with three random seeds to ensure statistical significance.

\subsection{Main Results}

\begin{figure*}[!h]
  \centering
  \includegraphics[width=1.00\linewidth]{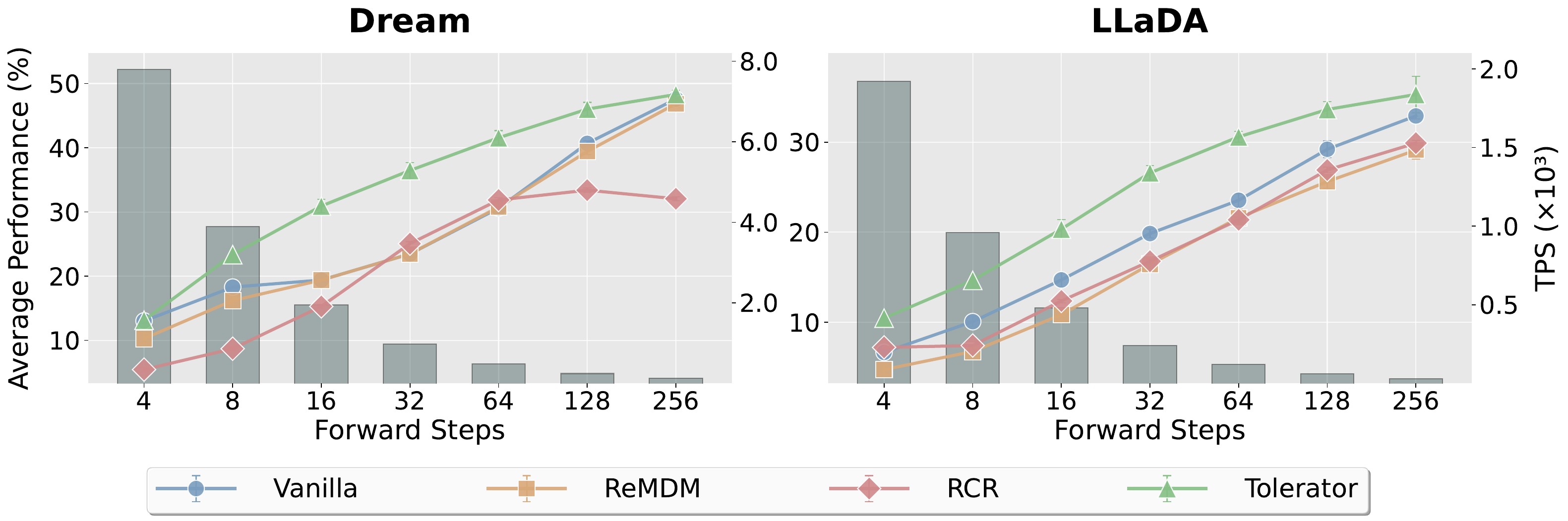}
  \vspace{-0.5em}
  \caption{\textbf{Performance-Efficiency Trade-Off for Different Decoding Methods.} This figure illustrates the performance of different methods under varying parallel sizes. Gray bars represent generation throughput (tokens per second, TPS). Colored lines show average performance across five benchmarks as forward step $T$ varies. }
  \label{fig: main_experiment_forward_steps}
  \vspace{-0.5em}
  
\end{figure*}
\begin{figure*}[!ht]
  \centering
  \includegraphics[width=1.00\linewidth]{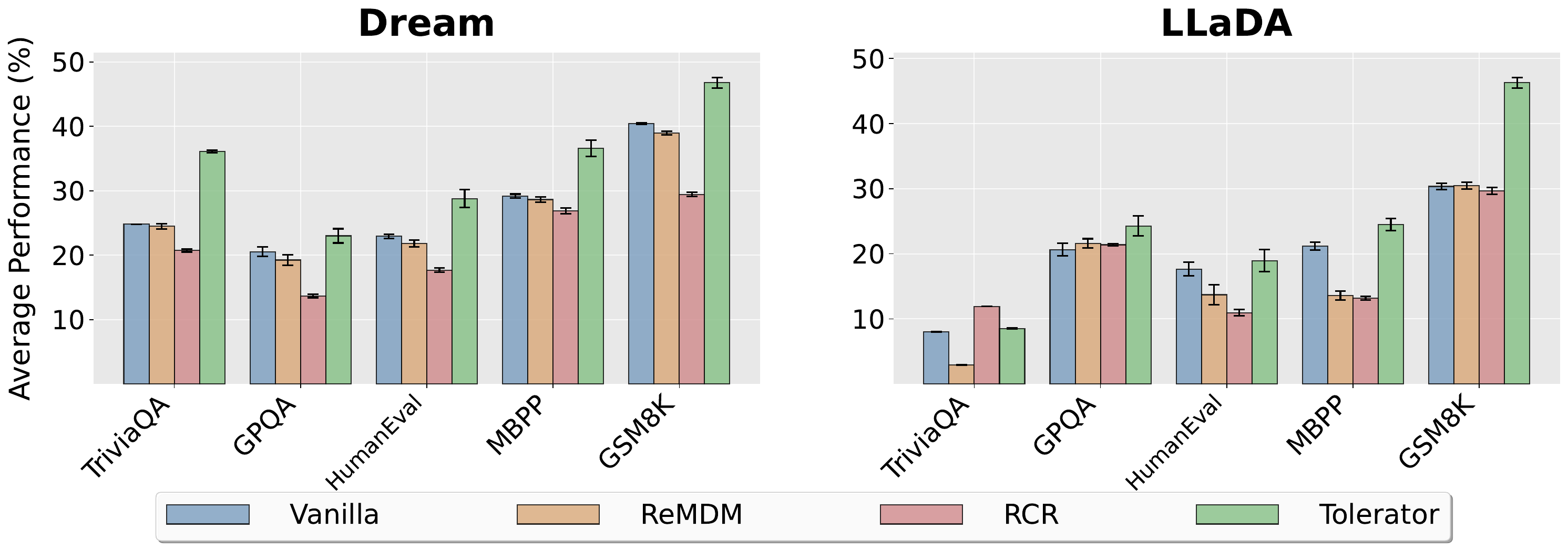}
  \vspace{-2.0em}
  \caption{\textbf{Performance across different benchmarks for different decoding methods. } This figure presents the performance of various methods under different benchmarks. Colored bars represent average performance across different forward steps ($T$).}
  \label{fig: main_experiment_tasks}
  \vspace{-1.5em}
\end{figure*}


To systematically evaluate performance across tasks and inference speeds, we compared our method against three baselines on five standard datasets. Some prior work~\citep{nie2025largelanguagediffusionmodels, ye2025dream7bdiffusionlarge, zhu2025llada} mainly focused on non-parallel decoding, where the number of forward steps equals the generation length. This setting achieves the highest accuracy but is also the most computationally expensive. In contrast, we additionally  examined parallel decoding scenarios, which we consider to be an essential feature of dLLMs and more common in practical applications~\citep{labs2025mercuryultrafastlanguagemodels, gemini_diffusion}. Specifically, for a generation length of $L = 256$, we varied the forward step budget $T$ over values of 4, 8, 16, 32, 64, 128, and 256. This corresponds to decoding 64, 32, 16, 8, 4, 2, and 1 tokens per forward pass. Figures~\ref{fig: main_experiment_forward_steps} and~\ref{fig: main_experiment_tasks} show the average performance across different forward steps and tasks (with detailed results in Appendix~\ref{app: experimental details}).

Figure~\ref{fig: main_experiment_forward_steps} illustrates the \emph{performance–efficiency} trade-off curves. \textsc{Tolerator} consistently improves generation quality across parallel decoding settings. On Dream, our method outperforms baselines with especially large gains at moderate parallelism levels (between 8 and 128 steps). On average, the percentage score increases from 29.0 to 34.6 (+17.9\% relatively). On LLaDA, improvements are similarly strong and remain stable even at extreme settings (steps = 4 and 256), with average performance increased from 21.3 to 24.5 (+15.3\% relatively). These results demonstrate that our approach improves over the baselines and adapts well across different levels of parallelism.

Figure~\ref{fig: main_experiment_tasks} summarizes performance across the five tasks, averaged over all forward steps. In most cases, our method outperforms the baselines. For instance, on Dream, the average score on TriviaQA improves from 24.8 to 36.1 (+45.16\% relatively), while on LLaDA, the score on GSM8K improves from 30.46 to 46.28 (+51.91\% relatively). Overall, these findings highlight the robustness and broad applicability of our method as a general improvement of dLLM decoding strategies.

\subsection{Ablation Studies}

 To analyze the effect of different components in our decoding strategy, we conduct ablation studies on (i) token-level cross-validation refinement, (ii) EoT Penalty, and (iii) refinement rate annealing.

\paragraph{Cross-Validation Refinement.}
\begin{table}[!ht]
\centering

\vspace{-1.5em}

\caption{\textbf{Performance under different refinement steps (\#$R$) with fixed fill-up stage steps.} Results are reported for both Dream-Instruct and LLaDA on GPQA and GSM8K.}

\vspace{1em}

\scalebox{0.75}{
\begin{tabular}{cllcccccccc}
\toprule
\textbf{Fill-Up} & \textbf{Model} & \textbf{Task} &
\#$R$=0 & \#$R$ = 4 & \#$R$ = 8 & \#$R$ = 16 & \#$R$ = 32 & \#$R$ = 64 & \#$R$ = 128 & \#$R$ = 256 \\
\midrule
\multirow{4}{*}{\textbf{Steps = 16}} 
 & Dream & GPQA   & 18.23 & 26.56 & 26.95 & 27.73 & 29.30 & 26.95 & 27.34 & 26.95 \\
 & Dream & GSM8K  & 26.79 & 41.41 & 42.19 & 47.66 & 58.20 & 64.45 & 65.23 & 66.80 \\
 & LLaDA & GPQA   & 19.87 & 25.39 & 22.66 & 25.39 & 21.09 & 23.83 & 24.22 & 24.61 \\
 & LLaDA & GSM8K  & 22.87 & 48.44 & 52.73 & 51.95 & 54.69 & 58.59 & 56.64 & 58.98 \\
\midrule
\multirow{4}{*}{\textbf{Steps = 64}} 
 & Dream & GPQA   & 22.25 & 28.12 & 31.64 & 31.64 & 31.64 & 35.16 & 35.55 & 30.08 \\
 & Dream & GSM8K  & 50.11 & 53.91 & 60.16 & 60.94 & 66.80 & 69.92 & 73.05 & 71.48 \\
 & LLaDA & GPQA   & 25.00 & 23.83 & 25.39 & 19.53 & 25.78 & 26.95 & 22.66 & 25.39 \\
 & LLaDA & GSM8K  & 40.99 & 59.38 & 64.06 & 66.02 & 67.97 & 65.23 & 71.09 & 69.14 \\
\bottomrule
\end{tabular}
}

\label{tab:refinement_effect}
\end{table}

We fix the number of forward steps in the fill-up stage at 16 and 64, then gradually increase the number of cross-validation refinement steps (\#$R$) from 0 to 256. The results are shown in Table~\ref{tab:refinement_effect}. We first observe that even a small number of refinement steps leads to substantial performance gains. For instance, on GSM8K, LLaDA achieves 22.87 with vanilla decoding in 16 steps. Adding just 4 refinement steps nearly doubles performance to 48.44.

Moreover, on GSM8K, increasing the number of refinement steps to 128 or even 256 continues to yield steady improvements. On GPQA, the trend is less pronounced, but the benefit of enabling cross-validation refinement is consistent: across all step counts, refinement outperforms the baseline without refinement (from 21.34 to 26.86 averaged over two models and two forward steps). Overall, these results demonstrate that the cross-validation refinement design is effective and well-motivated.

\paragraph{EoT penalty.}

\begin{figure*}[!ht]
  \centering
  \includegraphics[width=0.9\linewidth]{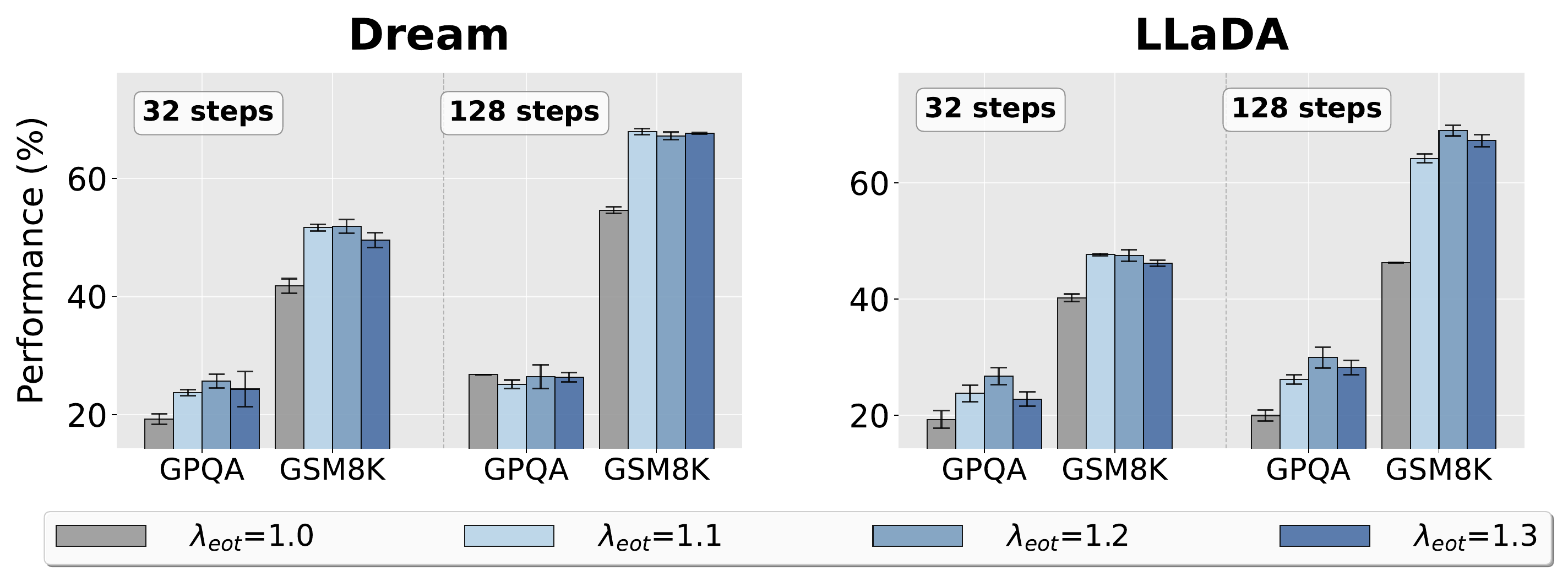}
  \vspace{-0.5em}
  \caption{\textbf{Ablation Studies of EoT Penalty.} We fix the fill-up and refinement configurations while varying $\lambda_{\text{eot}}$ from 1.0 to 1.3, with results shown for 32 and 128 forward step $T$. Across most tasks, introducing an appropriate EoT penalty substantially improves generation quality. The precise numerical values are reported in Appendix~\ref{app: experimental details}.}
  \label{fig: ablation_eot}
  \vspace{-0.5em}
  
\end{figure*}
To isolate the impact of the EoT penalty, we fix the fill-up and refinement configurations and vary only the penalty coefficient $\lambda_{\text{eot}}$. Specifically, we vary $\lambda_{\text{eot}}$ from 1.0 to 1.3 while keeping the number of forward step $T$ fixed at 32 and 128. 
We find that applying non-trivial $\lambda_{\text{eot}}$ consistently improves generation quality, with notable gains at $\lambda_{\text{eot}}=1.1$, $1.2$, and $1.3$ (+23.2\%, +28.4\%, +23.9\% relatively). 
This is because the EoT penalty encourages longer fill-up sequence: although these drafts may not always be fully correct, they tend to contain more information overall. During refinement, the useful content can be preserved and amplified while the incorrect parts are likely to be corrected.

\paragraph{Refinement Rate Annealing.}

\begin{table}[!h]
\centering
\vspace{-1em}
\caption{\textbf{Performance of LLaDA and Dream models with and without refinement rate annealing.} Across varying forward steps (\#$F$), annealing consistently improves accuracy for both models on GSM8K and TriviaQA.}
\vspace{1em}
\scalebox{0.85}{
\begin{tabular}{lllccccccc}
\toprule
\textbf{Model} & \textbf{Task} & \textbf{Setting} & \#$F$=4 & \#$F$=8 & \#$F$=16 & \#$F$=32 & \#$F$=64 & \#$F$=128 & \#$F$=256 \\
\midrule
\multirow{2}{*}{LLaDA} 
  & \multirow{2}{*}{GSM8K} & w/ Annealing    & 20.31 & 24.22 & 35.55 & 52.73 & 65.49 & 70.18 & 70.31 \\
  &                        & w/o Annealing   & 19.92 & 22.66 & 33.98 & 52.73 & 60.55 & 65.23 & 68.75 \\
\midrule
\multirow{2}{*}{Dream} 
  & \multirow{2}{*}{GSM8K} & w/ Annealing    & 12.37 & 25.26 & 36.20 & 50.00 & 62.76 & 69.53 & 74.09 \\
  &                        & w/o Annealing   & 14.06 & 23.05 & 33.98 & 47.27 & 64.84 & 67.97 & 70.31 \\
\midrule
\multirow{2}{*}{LLaDA}
  & \multirow{2}{*}{TriviaQA} & w/ Annealing   & 14.84 & 16.41 & 19.14 & 20.70 & 26.17 & 29.30 & 24.22 \\
  &                           & w/o Annealing  & 13.67 & 15.62 & 17.19 & 19.92 & 19.53 & 21.88 & 23.44 \\
\midrule
\multirow{2}{*}{Dream}
  & \multirow{2}{*}{TriviaQA} & w/ Annealing   & 0.78 & 0.78 & 1.56 & 3.52 & 8.98 & 12.50 & 16.02 \\
  &                           & w/o Annealing  & 0.39 & 2.34 & 1.17 & 0.78 & 4.30 & 10.94 & 15.23 \\
\bottomrule
\end{tabular}
}
\label{tab:annealing_effect}
\end{table}

In cross-validation refinement, we found that the ratio between sampled predicted tokens serving as validators and those serving as validation targets (referred to as the \emph{refinement rate} here) is critical. A rate that is too small makes it difficult to effectively revise validated tokens, while a rate that is too large limits the available context and lowers refinement quality. Thus, choosing an appropriate refinement rate is essential for stable performance. To achieve this balance, we introduce cosine annealing of the refinement rate, which provides sufficient momentum in the early steps of refinement while ensuring stability in the later steps.

To evaluate the effectiveness of this design, we compare the performance of Dream and LLaDA on GSM8K and TriviaQA with and without refinement rate annealing across different numbers of forward steps (Table~\ref{tab:annealing_effect}). For LLaDA, annealing consistently delivers an average improvement of +2.1 points across all step settings on GSM8K. For Dream, while exceptions occur at 4 and 64 steps, the overall effect is still an average improvement of +1.3 points on this dataset. On TriviaQA, refinement rate annealing remains effective across most forward steps, delivering average gains of +2.8 on LLaDA and +1.3 on Dream. These results demonstrate that refinement rate annealing is, overall, a robust and effective strategy.

Furthermore, Appendix~\ref{app: qualitative} presents several case studies that qualitatively illustrate the error correction process of our method.

\section{Discussion}






\subsection{Why our method can be training-free?}

Our method does not require additional training, yet still achieves significant improvements, because the refinement stage closely matches the training objective of dLLMs. During training, these models reconstruct randomly masked tokens given the visible context, and the loss is computed uniformly over all masked positions. This is exactly what our refinement stage does: in each iteration, a subset of tokens is remasked and predicted conditioned on the others, so that every token can in turn act as target or context under the same conditional distribution the model was optimized for.

Vanilla decoding, by contrast, is not fully consistent with this training setup. At inference it accepts a small set of the most confident tokens (based on logit features). The predictions are evaluated only on those high-confidence positions, while other masked tokens contribute nothing and simply get remasked—despite the fact that, during training, losses were computed on them as well.

\subsection{Why our strategy is good for large parallel sizes?}
Our method achieves better improvements in parallel decoding scenarios, i.e., when the forward step is smaller than the sequence length and multiple tokens are decoded in each forward step.

One key reason may lie in the visibility constraint during parallel decoding: tokens generated within the same step cannot attend to each other, which often leads to local inconsistencies. This phenomenon is even more noticeable with larger parallel sizes. Our token-level cross-validation process helps to mitigate this issue. During cross-validation, tokens filled up in the same step can be validated such that one serves as context (or validator) while another serves as the validation target. This mechanism enables tokens that were originally invisible to each other to interact directly—for example, when validating token A, token B (from the same step) can now be used as part of the context. Such interactions promote coherence among simultaneously decoded tokens. By repeating this process across multiple rounds, inconsistencies introduced by parallel decoding are progressively reduced, resulting in more coherent sequences overall.

In contrast, when the forward step equals the sequence length (i.e., non-parallel decoding with one token per step), every token naturally conditions on all previously accepted tokens. Since there is no within-step invisibility, the inconsistency problem does not arise, and thus the potential benefit of our method is relatively limited in this scenario.

\subsection{Limitations}

\paragraph{Format Stability.}

Although our method achieves consistent improvements across benchmarks, the gains are relatively smaller on code generation tasks such as HumanEval~\citep{chen2021codex} and MBPP~\citep{austin2021program}. These tasks are highly format-sensitive, where even slight deviations in syntax or structure can render an otherwise correct solution invalid. Because our refinement process operates purely at the token level and does not enforce explicit structural constraints, it can occasionally disrupt the formatting of well-formed code. This indicates a limitation of our approach when applied to domains that demand strict output adherence. A similar limitation has also been noted in methods such as RCR~\citep{he2025mdpoovercomingtraininginferencedivide}, which require heavier remasking than vanilla generation and can likewise compromise sequence formatting.

\paragraph{Natural Convergence.}

The cross-validation refinement in our method can be viewed as an iterative self-mapping process, where the input and output spaces are identical. Ideally, such a process should terminate naturally when, after a certain iteration, the sequence reaches a stable fixed point—subsequent updates would then map the sequence onto itself, eliminating the need to predefine a fixed number of steps. In practice, however, our current approach does not exhibit this kind of natural convergence. Even after many refinement steps, while the number of tokens edited by the model tends to decrease, it never converges to zero. Instead, the process continues to oscillate.

\section{Conclusion}

In this work, we revisited a key limitation of diffusion large language models (dLLMs): once a token is accepted during decoding, it is typically fixed and cannot be revised, causing early mistakes to persist and propagate through subsequent iterations. To address this, we proposed \textsc{Tolerator}, a training-free decoding strategy that explicitly decouples decoding into fill-up and refinement stages. By first generating a coarse draft and then iteratively remasking and decoding tokens with the token-level cross-validation principle, \textsc{Tolerator} enables more systematic and effective error correction than prior approaches.

Through extensive experiments on five benchmarks spanning natural language understanding, code generation, and mathematical reasoning, we showed that \textsc{Tolerator} consistently improves over baselines under the same forward step budgets. Beyond empirical gains, our results highlight that decoding strategy is not merely an implementation choice, but a crucial component that influences the overall performance of dLLMs.

\section*{Ethics Statement}

All datasets used in this work (TriviaQA~\citep{joshi-etal-2017-triviaqa}, GPQA~\citep{rein2024gpqa}, GSM8K~\citep{cobbe2021trainingverifierssolvemath}, HumanEval~\citep{chen2021codex}, MBPP~\citep{austin2021program}) are publicly available academic benchmarks that do not contain personally identifiable or sensitive information. Our study focuses on improving inference in discrete diffusion language models and does not involve the collection of new human subject data. We acknowledge that large language models may generate incorrect or misleading content, and that code generation models can potentially produce insecure or faulty programs. Our method does not eliminate these risks, and users should exercise caution when deploying such systems in high-stakes scenarios. The potential societal benefits of our work include improved decoding performance of diffusion large language models. This research was conducted in accordance with the ICLR Code of Ethics. The authors take full responsibility for all analyses and conclusions presented in this paper.

\section*{Reproducibility Statement}

We have taken several steps to ensure the reproducibility of our results. Our experiments were conducted on two representative open-source discrete diffusion language models: Dream-v0-Instruct-7B~\citep{ye2025dream7bdiffusionlarge} and LLaDA-8B-Instruct~\citep{nie2025largelanguagediffusionmodels}. We evaluate across five widely used public benchmarks—TriviaQA~\citep{joshi-etal-2017-triviaqa}, GPQA~\citep{rein2024gpqa}, GSM8K~\citep{cobbe2021trainingverifierssolvemath}, HumanEval~\citep{chen2021codex}, and MBPP~\citep{austin2021program}. For all methods, we adopt the same model backbones, zero-shot setting, and equalized computational budgets to guarantee fairness. Reported results are averaged over 3 random seeds, and exact numerical results for both main experiments and ablations are provided in the appendix. We detail hyperparameter configurations in Section 4.1, including scheduler settings, penalty coefficients, and baseline parameters (ReMDM~\citep{wang2025remaskingdiscretediffusionmodels} and RCR~\citep{he2025mdpoovercomingtraininginferencedivide}). Code, configuration files, and data preprocessing scripts are made anonymously available to facilitate replication. With the provided code and instructions, our results can be reproduced using 8×H200 GPUs or equivalent hardware.

\bibliography{references}
\bibliographystyle{iclr2026_conference}

\appendix

\section{Experimental Details}
\label{app: experimental details}

\subsection{Main Experiment}

\begin{table}[!h]
\centering
\caption{\textbf{Main Experiment Results}. Performance of Dream and LLaDA across five standard benchmarks under different numbers of forward steps. Highest values for specific task and model are \textbf{bold}. }
\label{tab: main results}
\vspace{0.5em}
\scalebox{0.72}{
\begin{tabular}{llccccccc}
\toprule

\multirow{2}{*}{\textbf{Model}} &
\multirow{2}{*}{\textbf{Method}} &
\multicolumn{7}{c}{\textbf{TriviaQA}} \\
\cmidrule(lr){3-9}
 & & \#F=4 & \#F=8 & \#F=16 & \#F=32 & \#F=64 & \#F=128 & \#F=256 \\
\midrule
\multirow{4}{*}{Dream}
 & Vanilla   & 23.08{\small $\pm$0.01} & 23.22{\small $\pm$0.02} & 23.16{\small $\pm$0.03} & 23.24{\small $\pm$0.02} & 23.51{\small $\pm$0.01} & 28.08{\small $\pm$0.03} & 29.32{\small $\pm$0.03} \\
 & ReMDM     & 22.11{\small $\pm$1.17} & 22.94{\small $\pm$0.28} & 22.87{\small $\pm$0.10} & 22.94{\small $\pm$0.16} & 23.27{\small $\pm$0.32} & 27.98{\small $\pm$0.41} & 29.26{\small $\pm$0.37} \\
 & RCR       & 15.63{\small $\pm$0.23} & 14.53{\small $\pm$0.12} & 15.02{\small $\pm$0.13} & 17.68{\small $\pm$0.13} & 18.92{\small $\pm$0.27} & 26.81{\small $\pm$0.34} & 36.64{\small $\pm$0.42} \\
 & \textsc{Tolerator} & \textbf{27.78}{\small $\pm$0.29} & \textbf{31.61}{\small $\pm$0.11} & \textbf{33.76}{\small $\pm$0.19} & \textbf{35.98}{\small $\pm$0.16} & \textbf{40.61}{\small $\pm$0.16} & \textbf{42.46}{\small $\pm$0.22} & \textbf{40.47}{\small $\pm$0.16} \\
\midrule

\multirow{4}{*}{LLaDA}
 & Vanilla   & 0.19{\small $\pm$0.02} & 0.65{\small $\pm$0.03} & 2.13{\small $\pm$0.01} & 4.63{\small $\pm$0.02} & 9.36{\small $\pm$0.06} & 16.25{\small $\pm$0.02} & 22.76{\small $\pm$0.01} \\
 & ReMDM     & 0.25{\small $\pm$0.02} & 0.43{\small $\pm$0.01} & 1.08{\small $\pm$0.02} & 1.82{\small $\pm$0.03} & 3.05{\small $\pm$0.06} & 5.43{\small $\pm$0.03} & 8.24{\small $\pm$0.02} \\
 & RCR       & 0.09{\small $\pm$0.01} & 0.80{\small $\pm$0.01} & \textbf{4.44}{\small $\pm$0.01} & \textbf{8.62}{\small $\pm$0.01} & \textbf{16.08}{\small $\pm$0.02} & \textbf{24.04}{\small $\pm$0.01} & \textbf{29.30}{\small $\pm$0.01} \\
 & \textsc{Tolerator} & \textbf{0.99}{\small $\pm$0.01} & \textbf{1.86}{\small $\pm$0.08} & 3.52{\small $\pm$0.09} & 6.19{\small $\pm$0.06} & 10.94{\small $\pm$0.10} & 16.46{\small $\pm$0.09} & 19.72{\small $\pm$0.14} \\

\midrule
\multirow{2}{*}{\textbf{Model}} &
\multirow{2}{*}{\textbf{Method}} &
\multicolumn{7}{c}{\textbf{GPQA}} \\
\cmidrule(lr){3-9}
 & & \#F=4 & \#F=8 & \#F=16 & \#F=32 & \#F=64 & \#F=128 & \#F=256 \\
\midrule
\multirow{4}{*}{Dream}
 & Vanilla   & \textbf{10.27}{\small $\pm$0.59} & 17.04{\small $\pm$0.34} & 18.23{\small $\pm$0.56} & 20.91{\small $\pm$1.10} & 22.25{\small $\pm$0.93} & 27.01{\small $\pm$0.80} & 27.98{\small $\pm$0.85} \\
 & ReMDM     & 7.44{\small $\pm$1.05} & 15.92{\small $\pm$1.03} & 17.93{\small $\pm$0.13} & 19.27{\small $\pm$0.90} & 22.62{\small $\pm$1.12} & 23.36{\small $\pm$0.72} & 28.20{\small $\pm$0.68} \\
 & RCR       & 1.79{\small $\pm$0.21} & 3.12{\small $\pm$0.13} & 7.81{\small $\pm$0.11} & 14.06{\small $\pm$0.28} & 25.00{\small $\pm$0.32} & 24.78{\small $\pm$0.41} & 19.20{\small $\pm$0.43} \\
 & \textsc{Tolerator} & 8.11{\small $\pm$1.05} & \textbf{17.19}{\small $\pm$0.97} & \textbf{22.84}{\small $\pm$0.13} & \textbf{26.71}{\small $\pm$1.45} & \textbf{26.93}{\small $\pm$1.10} & \textbf{29.91}{\small $\pm$1.77} & \textbf{29.32}{\small $\pm$1.23} \\
\midrule
\multirow{4}{*}{LLaDA}
 & Vanilla   & 10.79{\small $\pm$1.58} & 13.47{\small $\pm$1.58} & 19.87{\small $\pm$1.39} & 23.88{\small $\pm$0.67} & 25.00{\small $\pm$1.02} & 25.37{\small $\pm$0.46} & 26.04{\small $\pm$0.13} \\
 & ReMDM     & 9.60{\small $\pm$0.89} & 16.67{\small $\pm$0.13} & \textbf{23.66}{\small $\pm$1.18} & 24.70{\small $\pm$1.01} & 25.74{\small $\pm$0.68} & 25.82{\small $\pm$1.01} & 24.93{\small $\pm$0.13} \\
 & RCR       & 20.46{\small $\pm$0.46} & 18.45{\small $\pm$0.13} & 19.05{\small $\pm$0.13} & 18.97{\small $\pm$0.22} & 21.80{\small $\pm$0.13} & 26.19{\small $\pm$0.13} & 24.78{\small $\pm$0.13} \\
 & \textsc{Tolerator} & \textbf{20.76}{\small $\pm$1.46} & \textbf{20.76}{\small $\pm$1.46} & 22.47{\small $\pm$1.45} & \textbf{25.67}{\small $\pm$1.18} & \textbf{27.01}{\small $\pm$1.56} & \textbf{26.41}{\small $\pm$2.03} & \textbf{26.86}{\small $\pm$1.49} \\

\midrule
\multirow{2}{*}{\textbf{Model}} &
\multirow{2}{*}{\textbf{Method}} &
\multicolumn{7}{c}{\textbf{HumanEval}} \\
\cmidrule(lr){3-9}
 & & \#F=4 & \#F=8 & \#F=16 & \#F=32 & \#F=64 & \#F=128 & \#F=256 \\
\midrule
\multirow{4}{*}{Dream}
 & Vanilla   & \textbf{8.13}{\small $\pm$0.35} & 13.41{\small $\pm$0.00} & 11.79{\small $\pm$0.35} & 12.80{\small $\pm$0.61} & 26.02{\small $\pm$0.35} & 37.80{\small $\pm$0.61} & \textbf{50.61}{\small $\pm$0.00} \\
 & ReMDM     & 2.03{\small $\pm$0.70} & 9.35{\small $\pm$0.35} & 12.20{\small $\pm$0.00} & 13.21{\small $\pm$0.35} & 27.03{\small $\pm$0.70} & 38.82{\small $\pm$0.70} & 50.20{\small $\pm$0.93} \\
 & RCR       & 1.22{\small $\pm$0.24} & 8.54{\small $\pm$0.31} & 8.54{\small $\pm$0.31} & 22.56{\small $\pm$0.45} & 30.49{\small $\pm$0.37} & 26.22{\small $\pm$0.28} & 26.22{\small $\pm$0.28} \\
 & \textsc{Tolerator} & 4.88{\small $\pm$1.06} & \textbf{17.89}{\small $\pm$1.27} & \textbf{27.03}{\small $\pm$2.54} & \textbf{30.89}{\small $\pm$1.37} & \textbf{33.03}{\small $\pm$2.21} & \textbf{40.24}{\small $\pm$0.81} & 47.56{\small $\pm$0.61} \\
\midrule
\multirow{4}{*}{LLaDA}
 & Vanilla   & 9.55{\small $\pm$0.35} & \textbf{14.23}{\small $\pm$0.35} & 15.24{\small $\pm$1.22} & 15.45{\small $\pm$1.27} & 18.29{\small $\pm$1.40} & 23.68{\small $\pm$2.54} & \textbf{27.13}{\small $\pm$0.30} \\
 & ReMDM     & 4.88{\small $\pm$1.22} & 6.10{\small $\pm$0.00} & 8.13{\small $\pm$1.27} & 10.37{\small $\pm$1.22} & 18.09{\small $\pm$3.07} & 22.76{\small $\pm$0.70} & 25.61{\small $\pm$3.23} \\
 & RCR       & \textbf{9.96}{\small $\pm$0.35} & 5.08{\small $\pm$0.93} & 7.52{\small $\pm$0.35} & 7.93{\small $\pm$0.35} & 11.99{\small $\pm$0.35} & 15.85{\small $\pm$0.84} & 18.29{\small $\pm$0.31} \\
 & \textsc{Tolerator} & 7.52{\small $\pm$1.53} & 12.40{\small $\pm$0.35} & \textbf{20.43}{\small $\pm$1.40} & \textbf{23.58}{\small $\pm$0.93} & \textbf{22.05}{\small $\pm$0.77} & \textbf{24.19}{\small $\pm$0.77} & 22.46{\small $\pm$5.99} \\

\midrule
\multirow{2}{*}{\textbf{Model}} &
\multirow{2}{*}{\textbf{Method}} &
\multicolumn{7}{c}{\textbf{MBPP}} \\
\cmidrule(lr){3-9}
 & & \#F=4 & \#F=8 & \#F=16 & \#F=32 & \#F=64 & \#F=128 & \#F=256 \\
\midrule
\multirow{4}{*}{Dream}
 & Vanilla   & \textbf{14.40}{\small $\pm$0.20} & 14.73{\small $\pm$0.12} & 17.00{\small $\pm$0.20} & 25.00{\small $\pm$0.40} & 31.07{\small $\pm$0.12} & 45.13{\small $\pm$0.31} & \textbf{56.93}{\small $\pm$0.83} \\
 & ReMDM     & 8.80{\small $\pm$0.53} & 14.67{\small $\pm$0.31} & 15.93{\small $\pm$0.12} & 26.00{\small $\pm$0.20} & 33.13{\small $\pm$0.70} & 45.27{\small $\pm$0.64} & 56.60{\small $\pm$0.35} \\
 & RCR       & 4.80{\small $\pm$0.84} & 10.40{\small $\pm$0.71} & 23.60{\small $\pm$0.55} & 29.00{\small $\pm$0.29} & 36.00{\small $\pm$0.36} & 42.60{\small $\pm$0.43} & 41.73{\small $\pm$0.12} \\
 & \textsc{Tolerator} & 10.53{\small $\pm$1.01} & \textbf{25.13}{\small $\pm$0.64} & \textbf{35.00}{\small $\pm$0.80} & \textbf{41.07}{\small $\pm$2.91} & \textbf{44.40}{\small $\pm$1.20} & \textbf{48.47}{\small $\pm$1.55} & 51.53{\small $\pm$0.76} \\
\midrule
\multirow{4}{*}{LLaDA}
 & Vanilla   & \textbf{9.53}{\small $\pm$0.81} & 14.40{\small $\pm$0.40} & 13.40{\small $\pm$0.69} & 17.73{\small $\pm$0.12} & 24.07{\small $\pm$0.64} & 31.27{\small $\pm$0.81} & 37.87{\small $\pm$0.61} \\
 & ReMDM     & 1.53{\small $\pm$0.31} & 2.33{\small $\pm$0.58} & 4.53{\small $\pm$0.42} & 10.53{\small $\pm$0.12} & 17.27{\small $\pm$0.64} & 23.33{\small $\pm$1.72} & 35.47{\small $\pm$0.90} \\
 & RCR       & 0.60{\small $\pm$0.40} & 3.47{\small $\pm$0.46} & 10.00{\small $\pm$0.20} & 13.33{\small $\pm$0.12} & 15.93{\small $\pm$0.46} & 22.27{\small $\pm$0.31} & 26.73{\small $\pm$0.12} \\
 & \textsc{Tolerator} & 5.53{\small $\pm$1.03} & \textbf{16.00}{\small $\pm$0.87} & \textbf{22.73}{\small $\pm$1.03} & \textbf{25.60}{\small $\pm$0.69} & \textbf{29.27}{\small $\pm$0.42} & \textbf{33.87}{\small $\pm$0.81} & \textbf{38.53}{\small $\pm$1.50} \\

\midrule
 
\multirow{2}{*}{\textbf{Model}} &
\multirow{2}{*}{\textbf{Method}} &
\multicolumn{7}{c}{\textbf{GSM8K}} \\
\cmidrule(lr){3-9}
 & & \#F=4 & \#F=8 & \#F=16 & \#F=32 & \#F=64 & \#F=128 & \#F=256 \\
\midrule
\multirow{4}{*}{Dream}
 & Vanilla   & 9.22{\small $\pm$0.46} & 23.07{\small $\pm$0.04} & 26.79{\small $\pm$0.12} & 35.36{\small $\pm$0.09} & 50.11{\small $\pm$0.00} & 65.38{\small $\pm$0.12} & \textbf{73.10}{\small $\pm$0.06} \\
 & ReMDM     & 11.02{\small $\pm$0.64} & 18.12{\small $\pm$0.59} & 27.98{\small $\pm$0.00} & 35.91{\small $\pm$0.09} & 47.81{\small $\pm$0.09} & 61.66{\small $\pm$0.16} & 70.00{\small $\pm$0.44} \\
 & RCR       & 3.79{\small $\pm$0.25} & 6.90{\small $\pm$0.27} & 21.46{\small $\pm$0.33} & 42.00{\small $\pm$0.41} & 48.90{\small $\pm$0.38} & 46.55{\small $\pm$0.29} & 36.47{\small $\pm$0.35} \\
 & \textsc{Tolerator} & \textbf{14.40}{\small $\pm$0.59} & \textbf{24.92}{\small $\pm$1.22} & \textbf{35.96}{\small $\pm$1.45} & \textbf{47.66}{\small $\pm$0.18} & \textbf{62.80}{\small $\pm$0.90} & \textbf{68.99}{\small $\pm$0.92} & 72.61{\small $\pm$0.46} \\
\midrule
\multirow{4}{*}{LLaDA}
 & Vanilla   & 3.23{\small $\pm$0.18} & 7.58{\small $\pm$0.35} & 22.87{\small $\pm$0.70} & 37.55{\small $\pm$0.42} & 40.99{\small $\pm$0.61} & 49.46{\small $\pm$0.74} & 50.75{\small $\pm$0.57} \\
 & ReMDM     & 7.46{\small $\pm$0.16} & 8.24{\small $\pm$1.10} & 16.83{\small $\pm$0.40} & 34.77{\small $\pm$0.61} & 43.85{\small $\pm$0.24} & 50.77{\small $\pm$0.24} & 51.33{\small $\pm$0.81} \\
 & RCR       & 4.93{\small $\pm$0.32} & 9.29{\small $\pm$0.70} & 20.81{\small $\pm$0.27} & 35.03{\small $\pm$0.54} & 41.09{\small $\pm$0.64} & 46.17{\small $\pm$0.32} & 50.34{\small $\pm$0.96} \\
 & \textsc{Tolerator} & \textbf{17.49}{\small $\pm$0.43} & \textbf{22.24}{\small $\pm$0.90} & \textbf{32.58}{\small $\pm$1.20} & \textbf{51.88}{\small $\pm$1.14} & \textbf{63.66}{\small $\pm$0.23} & \textbf{67.20}{\small $\pm$0.64} & \textbf{68.89}{\small $\pm$1.05} \\

\bottomrule

\end{tabular}
}
\end{table}

In the main text, we present line and bar plots to highlight overall trends and comparisons on different tasks and forward step $T$. For completeness, here we report the exact numerical results of our main experiments in tabular form, which allow for more precise inspection and direct comparison across different methods and settings.

\FloatBarrier
\subsection{Ablation Studies}

Similarly, we present the exact numerical results our further analysis on EoT penalty in tables below.

\begin{table}[!ht]
\centering

\caption{\textbf{Performance with different values of the EoT penalty coefficient $\lambda_{\text{eot}}$ (1.0–1.3) under fixed fill-up and refinement configurations.} Evaluated on GPQA and GSM8K with Dream-Instruct and LLaDA. Reported as mean ({\small ± variance}) over 3 seeds.}

\vspace{1em}

\scalebox{1.0}{
\begin{tabular}{lllcccc}
\toprule
\textbf{Forward Steps} & \textbf{Model} & \textbf{Task} &
$\lambda_{\text{eot}}=1.0$ & 1.1 & 1.2 & 1.3 \\
\midrule
\multirow{4}{*}{\textbf{32}} 
 & Dream & GPQA   & 19.27 {\small ±0.90} & 23.74 {\small ±0.52} & 25.67 {\small ±1.18} & 24.33 {\small ±2.95} \\
 & Dream & GSM8K  & 41.80 {\small ±1.23} & 51.68 {\small ±0.57} & 51.88 {\small ±1.14} & 49.56 {\small ±1.25} \\
 & LLaDA & GPQA   & 19.27 {\small ±1.52} & 23.74 {\small ±1.44} & 26.71 {\small ±1.45} & 22.77 {\small ±1.24} \\
 & LLaDA & GSM8K  & 40.21 {\small ±0.64} & 47.66 {\small ±0.18} & 47.49 {\small ±1.01} & 46.17 {\small ±0.55} \\
\midrule
\multirow{4}{*}{\textbf{128}} 
 & Dream & GPQA   & 26.79 {\small ±0.00} & 25.15 {\small ±0.72} & 26.41 {\small ±2.03} & 26.34 {\small ±0.80} \\
 & Dream & GSM8K  & 54.61 {\small ±0.56} & 67.93 {\small ±0.53} & 67.20 {\small ±0.64} & 67.63 {\small ±0.13} \\
 & LLaDA & GPQA   & 19.94 {\small ±0.93} & 26.12 {\small ±0.80} & 29.91 {\small ±1.77} & 28.20 {\small ±1.23} \\
 & LLaDA & GSM8K  & 46.25 {\small ±0.08} & 64.22 {\small ±0.79} & 68.99 {\small ±0.92} & 67.27 {\small ±1.03} \\
\bottomrule
\end{tabular}
}

\label{tab:eot_penalty_effect}
\end{table}

\section{Use of LLMs Disclosure}

We disclose the following uses of large language models in the preparation of this work. GPT-5~\citep{openai2025gpt5} was employed solely to assist with language polishing and improving the readability of the manuscript. In addition, Claude Code~\citep{anthropic2025claudecode} was used as a coding assistant to generate and debug experimental scripts. At no point did LLMs contribute to the core research ideas, methodology, or interpretation of results. All scientific contributions, analyses, and conclusions remain the responsibility of the authors. Outputs produced by LLMs were carefully reviewed and revised where necessary to ensure accuracy and integrity.
\FloatBarrier 
\section{Qualitative Examples}

\label{app: qualitative}

In this section, we present a qualitative case study showing how cross-validation refinement performs token-level error correction. Consider the GSM8K~~\citep{cobbe2021trainingverifierssolvemath} query: “\textit{Marilyn's first record sold 10 times as many copies as Harald's. If they sold 88,000 copies combined, how many copies did Harald sell?}”

\begin{figure*}[!ht]
  
  \centering
  \includegraphics[width=0.7\linewidth]{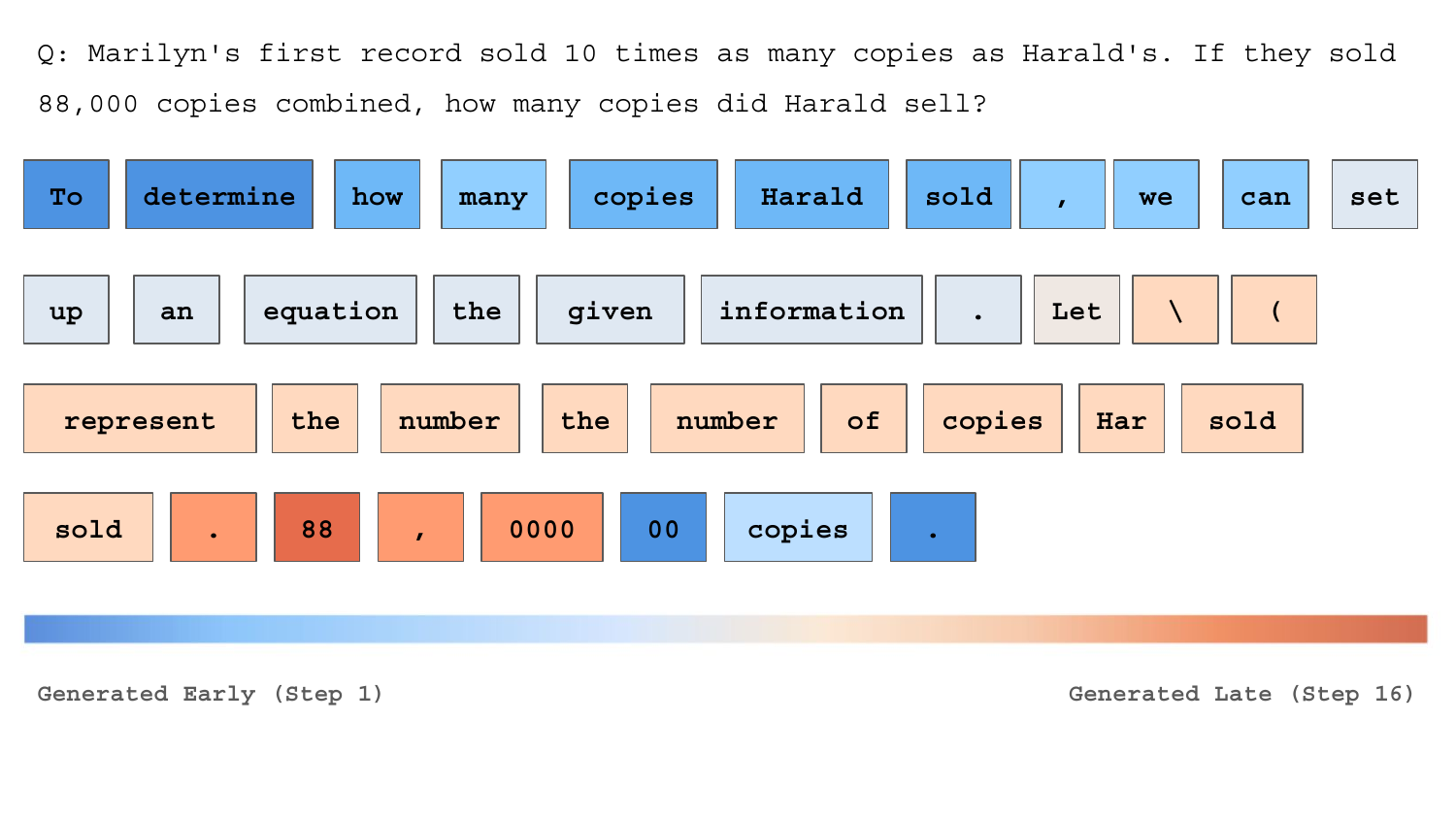}
  \vspace{-2.5em}
  \caption{\textbf{Output of Fill-Up Stage}.We use colors fading from blue to red to demonstrate the order of decoding.Using fill-up and refinement steps =16, special tokens like [EoT] are not shown.}
  \label{fig: qualitative_0}

  \centering
  \includegraphics[width=0.7\linewidth]{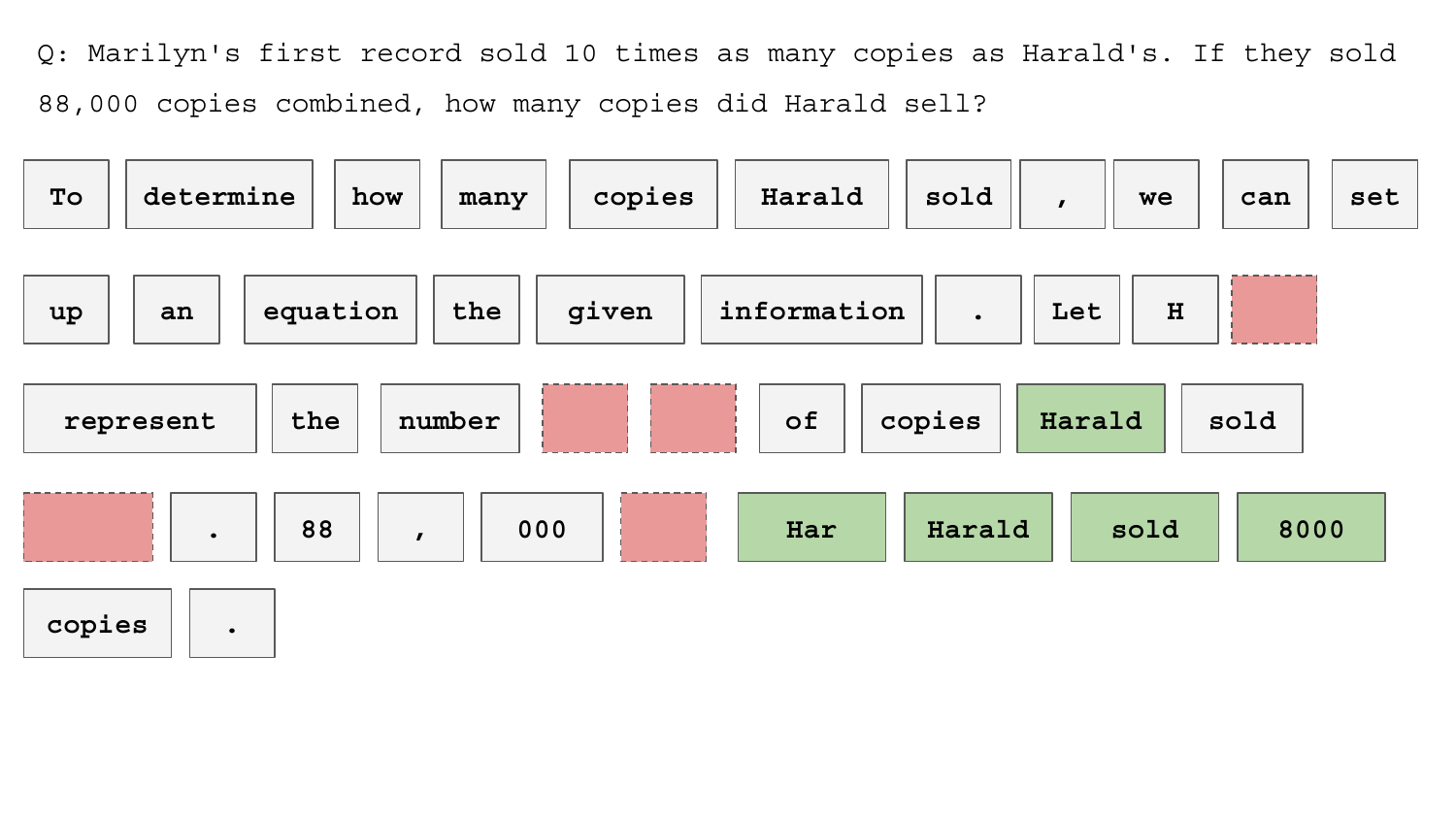}
  \vspace{-3.5em}
  \caption{\textbf{Sequence after 1 Iteration of Refinement. } Red dashed boxes represent deleted tokens while green boxes represent added tokens in current iteration.}
  \label{fig: qualitative_1}

  \centering
  \includegraphics[width=0.7\linewidth]{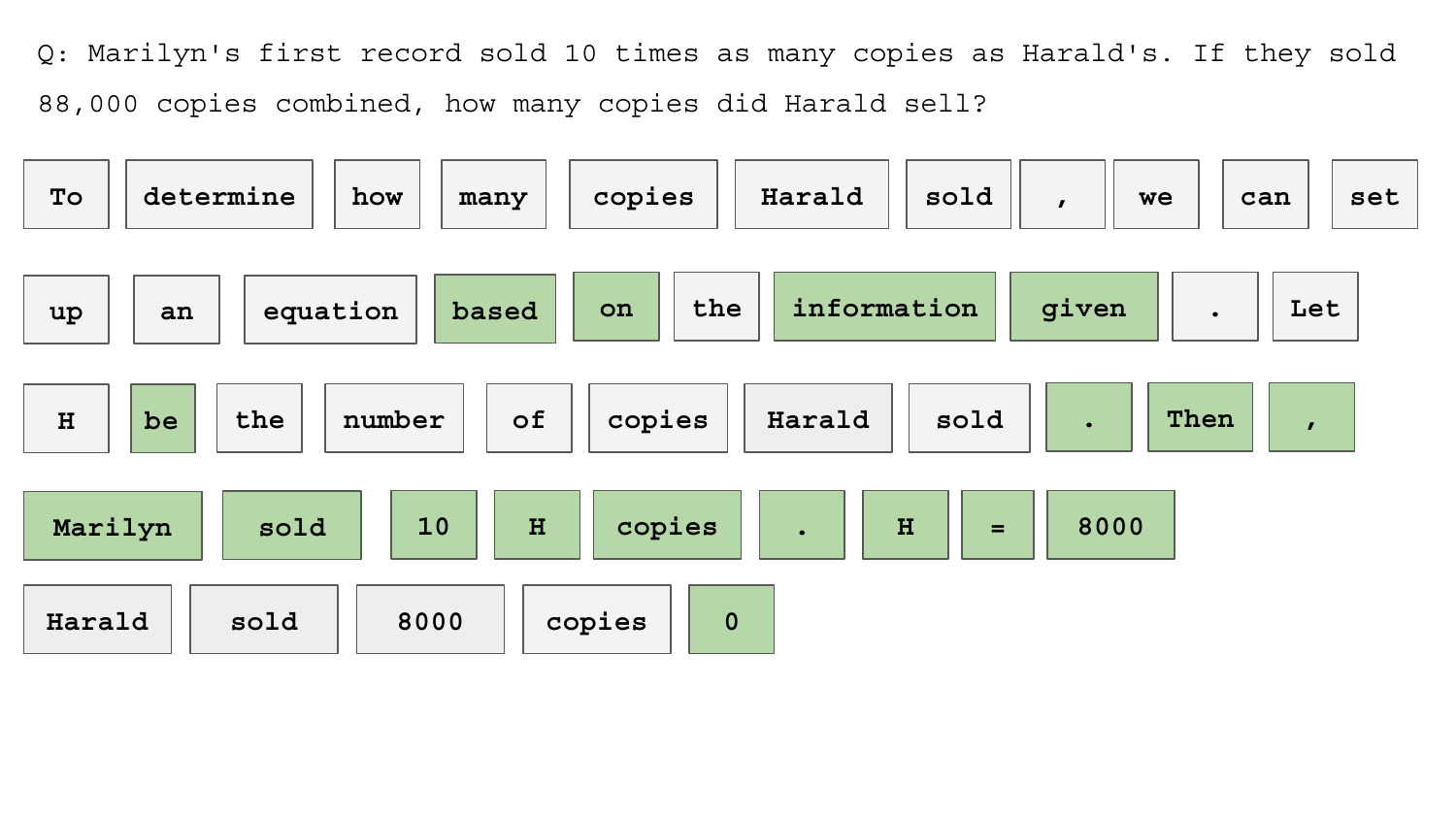}
  \vspace{-3.5em}
  \caption{\textbf{Sequence after 8 Iteration of Refinement}. Similarly, green boxes represent added tokens in current iteration. }
  \label{fig: qualitative_2}

  \centering
  \includegraphics[width=0.7\linewidth]{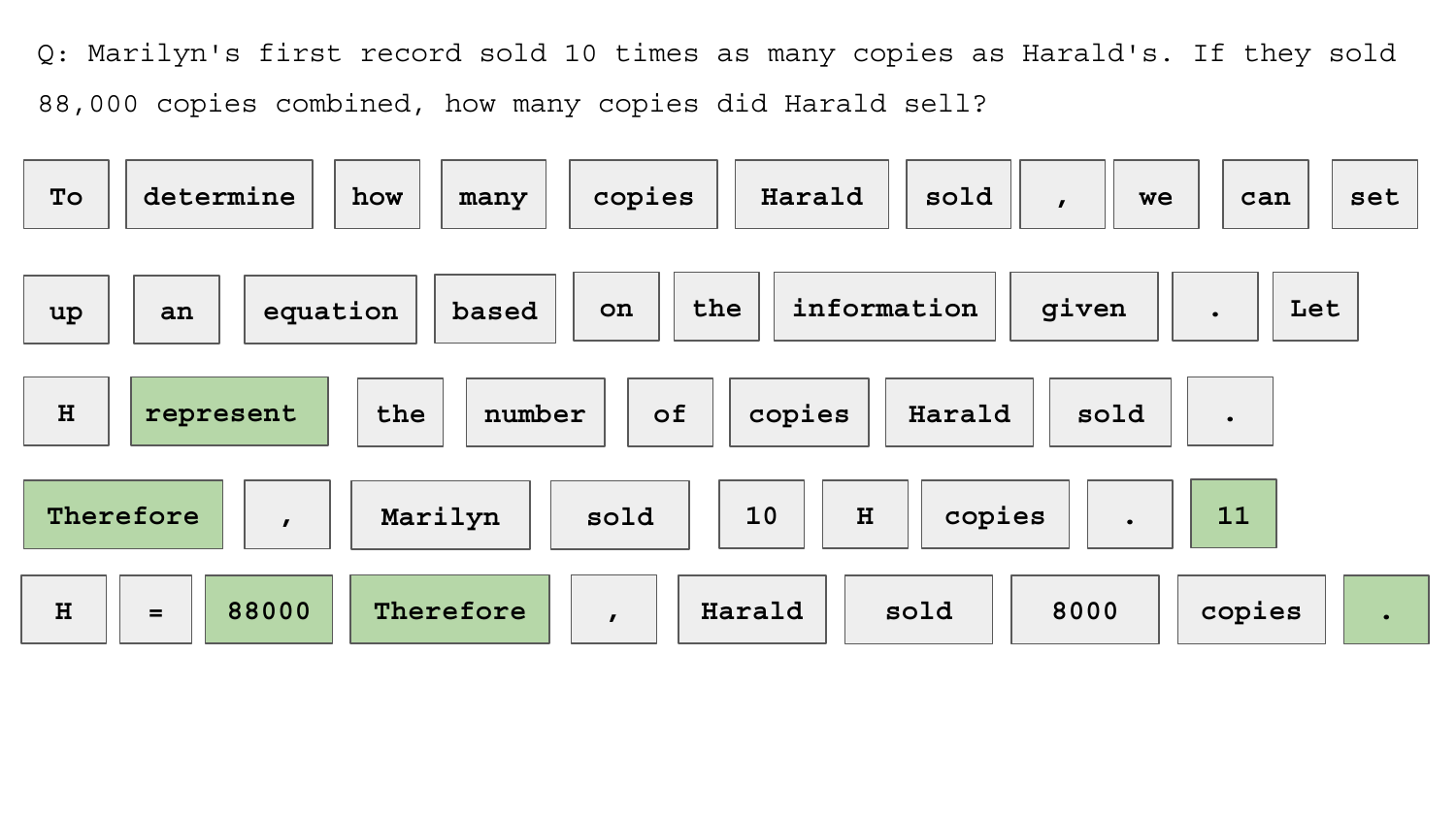}
  \vspace{-3.5em}
  \caption{\textbf{Sequence after 16 Iteration of Refinement}. Similarly, green boxes represent added tokens in current iteration. }
  \label{fig: qualitative_3}

\end{figure*}

Figure~\ref{fig: qualitative_0} shows LLaDA~\citep{nie2025largelanguagediffusionmodels} with 16 fill-up forward steps without refinement. Different colors indicate the decoding order. Early steps are syntactically and semantically reasonable, but by step 16 (orange) the output degrades—failing to carry out precise arithmetic (e.g., producing “88,000000”) and losing grammatical consistency (e.g., “the number the number”). 

Cross-validation refinement then iteratively edits inconsistent tokens and inserts more appropriate ones. Figures~\ref{fig: qualitative_1},~\ref{fig: qualitative_2} and~\ref{fig: qualitative_3} show results after 1, 8, and 16 refinement steps, respectively. After one step, redundant tokens like “the number the number” are removed and spurious digits are trimmed (e.g., “88,000000” → “88,000”), though minor issues (e.g., fragments like “Har Harald”) may remain. As refinement proceeds, the text becomes well-formed and the arithmetic is corrected, ultimately converging to the correct answer: 88,000.

This example indicates that dLLMs already contain the knowledge needed to solve the problem since initial generations include relevant fragments such as “88” and “00/0000”, but benefit from a more stable mechanism for parallel language modeling. Cross-validation refinement provides this mechanism, enabling correction over iterations.

\end{document}